\documentclass[10pt, a4paper]{article}
\usepackage{amsfonts}
\usepackage[final]{lrec2026} 

\usepackage{cleveref}

\usepackage{longtable}

\usepackage{multirow}
\usepackage{array}
\usepackage{booktabs}
\usepackage{booktabs,tabularx,xcolor}
\usepackage{booktabs}
\usepackage{siunitx}
\usepackage{subcaption}
\newcommand{\imp}{\(\uparrow\)}
\newcommand{\wor}{\(\downarrow\)}

\title{Towards Reward Modeling for AI Tutors \\in Math Mistake Remediation}

\name{Kseniia Petukhova, Ekaterina Kochmar} 

\address{Mohamed bin Zayed University of Artificial Intelligence \\
         \texttt{\{kseniia.petukhova, ekaterina.kochmar\}@mbzuai.ac.ae}\\}

\abstract{
Evaluating the pedagogical quality of AI tutors remains challenging: standard NLG metrics do not determine whether responses identify mistakes, scaffold reasoning, or avoid revealing the answers. For the task of mistake remediation, we derive a hierarchy of pedagogical aspects from human pairwise preferences on \texttt{MRBench}, and synthesize minimally contrastive response pairs that differ along key aspects (e.g., mistake identification and location, targetedness, scaffolding, actionability, clarity, and coherence). We develop and release Bradley-Terry preference models trained on weighted-sum rankings that we automatically create from \texttt{MRBench}, synthetic pairs, and data combinations. Using only synthetic data, our best model reaches 0.69 pairwise accuracy on a human preference test, and combining weighted-sum data with targeted synthetic groups improves accuracy to 0.74, outperforming larger general-purpose reward models while using only a 0.5B-parameter backbone.
\\ \newline \Keywords{AI tutoring, reward modeling, synthetic data}
}

\begin{document}

\maketitleabstract

\section{Introduction}

Human tutoring has long been recognized as one of the most effective forms of education, offering personalized guidance. Despite its success in improving learning outcomes \citep{bloom19842}, providing one-on-one tutoring at scale remains difficult due to the lack of qualified tutors and the high costs. Recent advances in large language models (LLMs) have opened new opportunities to address this gap. Their impressive conversational and reasoning abilities have led to a growth of interest in LLM-powered intelligent tutoring systems (ITSs) \citep{pal2024autotutor,liu2024teaching}. These systems can assist learners across diverse educational tasks \citep{wollny2021we}, among which supporting students in understanding and remediating their mistakes has become a particularly promising application \citep{macina2023mathdial,wang2023bridging}.

While substantial progress has been made in building AI tutoring systems, evaluating their pedagogical quality remains a significant challenge. Conventional natural language generation (NLG) metrics \citep{chin2004rouge,popovic2017chrf++,post2018call,gao2020dialogue,liu2023g} cannot be used for this purpose, as they are domain-agnostic, rely on reference texts, and fail to capture the nuanced pedagogical aspects of effective tutoring. In the context of student mistake remediation, an ideal AI tutor should guide the learner toward understanding rather than simply reveal the answer, while maintaining factual accuracy, clarity, and a supportive tone. However, studies show that existing LLMs often fall short, simply revealing solutions without providing sufficient guidance or explanation \citep{macina2023mathdial}.

\begin{figure}[!t]
  \centering
  \includegraphics[width=0.95\linewidth]{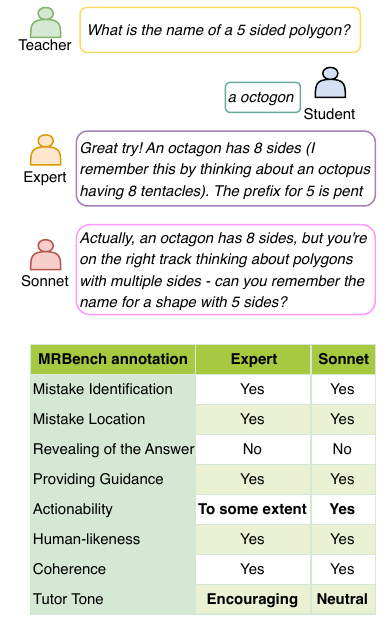}
  \caption{Example of annotated tutor responses from \texttt{MRBench}. The \texttt{Sonnet} response is annotated as more actionable (``Yes'' vs. ``To some extent'') because it prompts the student to recall the prefix for a five-sided shape rather than revealing it directly. In contrast, the \texttt{Expert} response is more encouraging in tone (``Encouraging'' vs. ``Neutral'').}
  \label{fig:octagon}
\end{figure}

To facilitate systematic evaluation of AI tutors, \citet{maurya2024unifying} introduced a taxonomy for assessing pedagogical abilities of LLM-powered tutors and released \texttt{MRBench} -- a benchmark containing tutor responses annotated across eight pedagogical dimensions: \textit{mistake identification} (whether the tutor identifies an error), \textit{mistake location} (whether it pinpoints where the error occurs), \textit{providing guidance} (whether the tutor provides relevant guidance), \textit{actionability} (whether it is clear from the response what to do next), \textit{revealing of the answer}, \textit{coherence}, \textit{tutor tone}, and \textit{human-likeness}. Each dialog context in the dataset contains 8-9 alternative tutor responses produced by humans or LLMs. Building on this work, \citet{kochmar2025findings} organized a shared task for automatic evaluation across the first four dimensions. Despite promising results on these individual aspects, the question of determining the overall quality of a tutor response remains open. For example, one response might excel at \textit{mistake location}, while another is more \textit{actionable}. Prior work, such as \citet{macina2025mathtutorbench}, addressed this by assigning equal weights to all dimensions and summing the scores. However, this approach leads to frequent ties, and it assumes uniform importance across pedagogical aspects -- an unrealistic assumption, since dimensions like \textit{providing guidance} likely carry greater pedagogical weight than \textit{tutor tone} or \textit{human-likeness}. An example of such a tie is shown in \Cref{fig:octagon}, where two responses receive identical overall scores. The \texttt{Sonnet} response achieves a full score for \textit{actionability}, whereas \texttt{Expert} receives only a partial score; conversely, \texttt{Expert} receives a full score for \textit{tutor tone}, while \texttt{Sonnet} does not. Despite having the same total score, the responses differ in pedagogical quality: the \texttt{Sonnet} response is more actionable, prompting the student to recall the correct prefix, while the \texttt{Expert} response is more encouraging but partially reveals the answer.

To address these limitations, we conduct human pairwise preference annotation to identify which pedagogical aspects most strongly influence judgments of response quality. Using these insights, we construct a synthetically generated dataset of response pairs that differ along these key aspects and train a ranking model that captures human preferences. The dataset and the trained model are publicly released to support future research on pedagogical evaluation and reward modeling for AI tutors. This model can serve as an evaluation tool or reward model for aligning LLM-based tutors toward improved pedagogical behavior.


In summary, our key contributions are as follows:\vspace{-0.5em}
\begin{itemize}
    \item We perform human pairwise preference annotation on responses from \texttt{MRBench}~\cite{maurya2024unifying} and derive a hierarchy of pedagogical aspects that humans prioritize when selecting the better response.\vspace{-0.5em}
    \item We propose a method for synthetically generating response pairs that differ systematically along these aspects, creating a new dataset of pedagogically contrastive tutor responses.\vspace{-0.5em}
    \item We train a reward model that outperforms existing general-purpose reward models, making it better suited for evaluating and aligning LLM-powered tutoring systems.\vspace{-0.5em}
    \item We release all resources developed in this work, including the annotated human preference data, the synthetic augmentation dataset, the trained reward model, and the code, to support further development of pedagogically-aligned AI tutors: \url{https://github.com/Kpetyxova/Towards_Reward_Modeling_for_Tutors}\vspace{-0.5em}
    \item Although our experiments focus on math mistake remediation, the proposed methodology for preference data generation can be applied to other domains.
\end{itemize}

\section{Related Work}

\paragraph{Pedagogical Ability Assessment} \citet{maurya2024unifying} introduce a systematic framework to evaluate the pedagogical capabilities of LLM-powered tutors. The main contribution of this paper lies in its creation of the first unified evaluation taxonomy explicitly designed for assessing AI tutors' pedagogical abilities in student mistake remediation tasks. Prior research evaluated tutoring models using fragmented and inconsistent criteria, making cross-model comparison difficult. This work unifies these diverse efforts into a coherent framework of eight pedagogical dimensions: \textit{mistake identification}, \textit{mistake location}, \textit{revealing of the answer}, \textit{providing guidance}, \textit{actionability}, \textit{coherence}, \textit{tutor tone}, and \textit{human-likeness}, grounded in learning sciences principles. For the \textit{revealing of the answer} dimension, the labels are \underline{\textit{No}}, \textit{Yes (correct answer)}, and \textit{Yes (incorrect answer)}. For the \textit{tutor tone}, the possible values are \textit{Negative}, \textit{Neutral}, and \underline{\textit{Encouraging}}. Each remaining dimension can take one of three possible values: \underline{\textit{Yes}}, \textit{To some extent}, or \textit{No}.\footnote{\underline{Underlined} are the preferred options.}

Alongside the taxonomy, the authors introduce \texttt{MRBench}, a new benchmark of 192 dialogs and 1,596 tutor responses (human and LLM-generated), annotated with human gold labels. \texttt{MRBench} is constructed from contexts and tutor responses drawn from the \texttt{MathDial} \citep{macina2023mathdial} and \texttt{Bridge} \citep{wang2023bridging} datasets and extended with responses generated by seven state-of-the-art LLMs acting as tutors. Human evaluation of these responses reveals that while linguistically fluent, current LLMs often lack pedagogical depth. Models such as \texttt{GPT-4} and \texttt{Llama-3.1-405B} perform well in identifying and locating student mistakes but tend to reveal answers too frequently, which diminishes their instructional effectiveness. \texttt{Mistral} and \texttt{Sonnet} show moderate performance, whereas \texttt{Gemini} and \texttt{Phi-3} perform poorly, producing incoherent or unhelpful guidance. Smaller models like \texttt{Llama-3.1-8B} achieve relatively balanced yet still suboptimal results. Finally, the authors explore whether LLMs can serve as evaluators across the proposed dimensions. Correlation analyses show that LLM-based evaluators remain unreliable, as their judgments correlate weakly or negatively with human assessments across most pedagogical dimensions.

Following this, \citet{kochmar2025findings} present the outcomes of a shared task aimed at automating the evaluation of AI tutors' pedagogical abilities across four of the above dimensions: \textit{mistake identification}, \textit{mistake location}, \textit{providing guidance}, and \textit{actionability}. Over 50 international teams participated, employing diverse techniques including LoRA-based fine-tuning, in-context learning, data augmentation, ensemble modeling, and reinforcement learning with human feedback. The results demonstrate meaningful progress but highlight persistent pedagogical quality assessment challenges. Across the four tracks, the best-performing systems achieved macro F1 scores ranging from 58.3 (for \textit{providing guidance}) to 71.8 (for \textit{mistake identification}), indicating that automated systems can capture some pedagogical dimensions but still fall short of human-level understanding.

In contrast to the shared task approach, \citet{macina2025mathtutorbench} train a reward model that scores the overall pedagogical quality of generated tutor utterances. This model is trained using pairwise ranking on preference data derived from expert and novice teacher responses, and from annotations from \texttt{MRBench} -- for each response, a cumulative score is computed by summing the presence of desirable values across eight evaluation dimensions. The resulting reward model achieves an accuracy of 0.84 in distinguishing expert from novice teacher responses. However, this approach often produces tied scores and implicitly assumes equal importance across all eight dimensions. Furthermore, the evaluation test set may overestimate performance, as it is based on the \texttt{Bridge} dataset, where novice responses are typically much weaker than those of expert tutors and are characterized by repetitive, generic feedback patterns such as ``try again'' or ``check your answer.''

\paragraph{Generation of Synthetic Preference Pairs} To enhance the quality of preference data used for reward-model training, synthetic preference generation techniques have been developed. For example, \citet{pace2024west} propose a method in which responses are generated in pools and then the best and worst are paired to create synthetic preference pairs. \citet{shen2024boosting} present another approach: they first generate one response, assign a desired preference label, and then generate a second response conditioned on the first one plus the label and guidance about multiple quality aspects. At the same time, Contrastive Learning from AI Revisions (CLAIR) \citep{d2025anchored} takes a different approach: it produces minimally contrastive preference pairs by having a stronger model revise the weaker model's output rather than selecting between two independently generated outputs. In CLAIR, the target model generates an initial losing output for each prompt, which a stronger reviser model then minimally improves to create the winning output. This ensures that differences between the two outputs are targeted and relevant, reducing noise from irrelevant differences. Their results show that CLAIR produces preference pairs with the highest measured contrastiveness and yields the most significant performance gains when combined with APO-zero (an alignment objective proposed in the same work). Inspired by CLAIR's minimal-revision paradigm and related multi-aspect synthetic preference generation methods such as the one from \citet{shen2024boosting}, we transfer these ideas to the educational domain, generating targeted revisions of tutor responses along key valuable aspects.

\section{Creation of Preference Pairs}

\subsection{Preliminary Ranking}
\label{sec:preliminary_ranking}

In this section, we construct a preliminary set of preference pairs following the approach of \citet{macina2025mathtutorbench}. However, we modify the weighting scheme to account for the varying importance of different evaluation dimensions. We hypothesize that some pedagogical aspects contribute more strongly to human judgments of response quality than others, and we use this assumption as a starting point for our analysis. Specifically, the most important dimensions are hypothesized to be \textit{mistake location}, \textit{providing guidance}, and \textit{coherence}; moderately important dimensions include \textit{mistake identification} and \textit{actionability}; less important dimensions are \textit{revealing the answer}, whose importance is context-dependent (early in the dialog revealing the answer is undesirable, but when the student is clearly stuck, it becomes necessary), and \textit{human-likeness}; and, comparatively, the least important of eight dimensions is \textit{tutor tone}.\footnote{This hypothesis is based on manual inspection of the dialogs and tutor responses quality. We highlight that it is only used to produce the rankings of the responses as the starting point.}

To calculate the overall score for each response, we first map each annotation label to a numerical value. For most dimensions, we assign a value of 1 to \textit{Yes}, 0.5 to \textit{To some extent}, and 0 to \textit{No}. For \textit{Revealing of the Answer}, the scoring is asymmetric: 1 if the tutor does not reveal the answer, 0.5 if the tutor reveals the correct answer, and 0 if the tutor reveals an incorrect answer. For \textit{Tutor Tone}, we assign 1 to Encouraging, 0.5 to Neutral, and 0 to Offensive. Each numerical annotation score is then multiplied by the corresponding dimension weight: \textit{Mistake Identification} (0.5), \textit{Mistake Location} (1.0), \textit{Revealing of the Answer} (0.25), \textit{Providing Guidance} (1.0), \textit{Actionability} (0.5), \textit{Human-likeness} (0.25), \textit{Coherence} (1.0), and \textit{Tutor Tone} (0.05). The resulting weighted sum serves as an estimate of the pedagogical quality of a tutor's response. Across the 1,655 responses in the dataset, the resulting scores range from 0.025 to 4.55 (mean = 3.45, median = 3.90, SD = 1.28). The interquartile range spans from 2.78 (25th percentile) to 4.53 (75th percentile), indicating that most responses receive relatively high scores, with a moderate spread across the scale. This distribution suggests that while many responses exhibit strong pedagogical qualities, meaningful variation remains, allowing for comparative ranking. Overall, this approach provides an initial quantitative basis for ranking responses by their instructional effectiveness.

To evaluate the preliminary ranking, we conducted a human evaluation study involving four annotators, each holding at least a Master's degree in NLP or Computer Science.  Each annotator evaluated tutor responses drawn from ten dialogs, five of which were shared among all participants. For every dialog, annotators assessed all response pairs with a score difference less than or equal to 0.5 -- on average, about ten such pairs per dialog. In addition, each annotator evaluated three to five pairs per dialog where the score difference exceeded 0.5. The underlying intuition was that larger score differences could generally be trusted as more reliable, whereas pairs with minor or tied differences required closer human inspection. For instance, the responses shown in \Cref{fig:octagon} represent a pair with a score difference below 0.5, illustrating a slight contrast. The annotators labeled 152, 135, 155, and 153 pairs, respectively.

Across the five shared dialogs, there were 86 pairs: 64 with a score difference of 0.5 or less and 22 with a greater difference. For each pair, annotators were asked to select which response was better or indicate that both were either good or poor. After the initial annotation round, a group discussion was held to establish \textbf{hierarchy of pedagogical aspects} (see Appendix \ref{ap:hierarchy}), summarized as follows:
\begin{enumerate}
    \item \textit{Factuality + Non-contradiction + No Nonsense}: The response should be factually correct, should not contradict the student's answer(s), and should not contain irrelevant information. Two responses are considered equally poor when such aspects are lacking in both.
    \item \textit{Mistake Identification + Location}: The response should explicitly or implicitly identify a mistake in the student's solution. For example, saying ``Nice try'' would miss this aspect.
    \item \textit{Scaffolding + Actionability}: The response should address the misunderstanding or problem step by step, guiding the student toward understanding by asking questions or providing hints instead of directly giving away the solution. However, providing the solution is preferable if the context shows that scaffolding has not worked.
    \item \textit{Targetedness}: The response should address the core misconception or misunderstanding of a student.
    \item \textit{Not revealing the final answer}: While sharing the answer to a substep is sometimes necessary and acceptable, the tutor should avoid giving away the final answer.
    \item \textit{Clarity + Coherence}: The tutor's response should be free of awkward, confusing, or misleading wording. A good tutor response acknowledges the student's input and connects it to the next step.
\end{enumerate}

The inter-annotator agreement among the four annotators, measured using Fleiss' Kappa across all shared pairs, was 0.72, indicating substantial agreement. For shared pairs with a score difference greater than 0.5, annotators achieved perfect agreement (Fleiss' Kappa = 1.0) and agreed with the preliminary ranking in 88\% of cases. This outcome is expected, as pairs with a score difference above 0.5 represent responses that are reliably distinct in quality, making them easier to judge consistently. The agreement was lower for pairs with minor score differences or ties, with a Fleiss' Kappa of 0.62. An illustrative case of annotator disagreement for such borderline pairs is provided in Appendix~\ref{ap:annotator_disagreement}, where differing pedagogical preferences led to divergent judgments despite comparable overall quality.

\Cref{tab:misalignment_1} shows an example: human annotators preferred the second response because it provided scaffolding, which they prioritized. However, according to the \texttt{MRBench} annotation, this response scored worse, since it only partially identified the mistake and did not locate it. Additional examples of such discrepancies are presented in Appendix~\ref{ap:misalignment_examples}. Together, these cases highlight that perfect alignment between the weighting scheme and human annotation is unattainable with the current \texttt{MRBench} dimensions. For instance, \textit{providing guidance} and \textit{actionability} do not distinguish between scaffolding and more detailed explanations; a \textit{providing guidance} label of \textit{No} could mean either that guidance is missing or that guidance is present but factually incorrect. Consequently, responses that prioritize scaffolding may receive lower scores on \textit{mistake identification} and \textit{mistake location}, revealing a structural tension between these dimensions.

\begin{table*}[!htb]
\centering
\small
\resizebox{\linewidth}{!}{
\begin{tabular}{|
  >{\raggedright\arraybackslash}p{0.10\linewidth}|
  >{\raggedright\arraybackslash}p{0.28\linewidth}|
  c|c|c|c|c|c|c|c|c|}
\hline
\multicolumn{11}{|l|}{\textbf{Dialog Context}} \\ \hline
\multicolumn{11}{|p{\linewidth}|}{
\textbf{Student:} yay

\textbf{Tutor:} Next, we have to find the value of 6 times 9 + 2.

\textbf{Tutor:} What is the value of 6 times 9 + 2?

\textbf{Student:} 99
} \\ \hline\hline

\textbf{Preferred Response} & \textbf{Tutor Response}
& \textbf{MI} & \textbf{ML} & \textbf{RA}
& \textbf{PG} & \textbf{AC} & \textbf{HM}
& \textbf{CO} & \textbf{TT} & \textbf{Score} \\
\hline

\multirow[t]{2}{0.1\linewidth}{Preferred (weighted-sum)}
& Good try, but remember the order of operations: multiplication should be done
before addition. So it should be 6 times 9, then add 2. Let's try again!
& Yes & Yes & No & Yes & Yes & Yes & Yes & Enc. & \multirow{2}{*}{4.55} \\
\cline{3-10}
& 
& 0.50 & 1.00 & 0.25 & 1.00 & 0.50 & 0.25 & 1.00 & 0.05 & \\
\hline

\multirow[t]{2}{0.1\linewidth}{Preferred (human)}
& Great try! What was your first step?
& TSE & No & No & Yes & Yes & Yes & Yes & Enc. & \multirow{2}{*}{3.30} \\
\cline{3-10}
&
& 0.25 & 0.00 & 0.25 & 1.00 & 0.50 & 0.25 & 1.00 & 0.05 & \\
\hline

\end{tabular}
}
\caption{Example of misalignment between the weighted-sum ranking (Section~\ref{sec:preliminary_ranking}) and human preference. For each response, the first row reports the original \texttt{MRBench} annotation labels, and the second row reports their weighted contributions (annotation score $\times$ dimension weight). Abbreviations: MI -- \textit{Mistake Identification}; ML -- \textit{Mistake Location}; RA -- \textit{Revealing the Answer}; 
PG -- \textit{Providing Guidance}; AC -- \textit{Actionability}; HM -- \textit{Human-likeness}; CO -- \textit{Coherence}; TT -- \textit{Tutor Tone}; TSE -- \textit{To some extent}; Enc -- \textit{Encouraging}.}
\label{tab:misalignment_1}
\end{table*}

Thus, while the proposed preliminary ranking provides a reasonable approximation and can be used as a silver standard, it cannot fully capture the nuanced preferences observed during human annotation. A promising direction is to use the annotated data and the hierarchy of valuable aspects to prompt LLMs to detect such misalignments -- our next step in this work.

\subsection{Can We Use LLMs for Preference Annotation?}
\label{sec:llm_annotation}

Since the proposed preliminary ranking cannot serve as a gold standard, we examine whether LLMs can be used for preference annotation. We tested four prompting strategies -- {\em basic}, {\em with guidelines}, {\em with hierarchy}, and {\em with checklist}, across three proprietary models: \texttt{GPT-4.1}, \texttt{Claude-sonnet-4}, and \texttt{Gemini-1.5-pro}. The full prompts are provided in Appendix~\ref{ap:prompts}.

Evaluation on 337 human-annotated pairs showed accuracies between 0.58 and 0.66 across models. The {\em with guidelines} prompt performed best (0.66 for \texttt{Claude-sonnet-4} and \texttt{Gemini-1.5-pro}; 0.64 for \texttt{GPT-4.1}), and an ensemble of all models reached 0.67. Although moderately aligned with human judgments, these results remain insufficient for reliable large-scale annotation.

\subsection{Synthetic Data Augmentation}

Building on the concept of minimal revisions proposed by \citet{d2025anchored}, we introduce a synthetic data augmentation procedure designed to enrich \texttt{MRBench} with additional preference pairs that vary meaningfully along the aspects defined during the preference annotation in the hierarchy of pedagogical aspects. 

In designing this procedure, we intentionally focus on the \textit{revealing the answer}, \textit{providing guidance}, \textit{actionability}, and \textit{coherence} dimensions from \texttt{MRBench} when grouping the original responses for each step we describe below. As discussed in the misalignment example in \Cref{tab:misalignment_1}, low scores in the \textit{Mistake Identification} (\textit{MI}) and \textit{Mistake Location} (\textit{ML}) dimensions do not necessarily indicate poor pedagogical quality -- some high-quality responses may provide effective scaffolding without explicitly identifying the mistake. To avoid penalizing such cases, we exclude the \textit{MI} and \textit{ML} dimensions from the filtering process and instead prioritize aspects that more directly capture instructional quality and student understanding.

\begin{figure*}[!ht]
  \centering
  \includegraphics[width=0.95\linewidth]{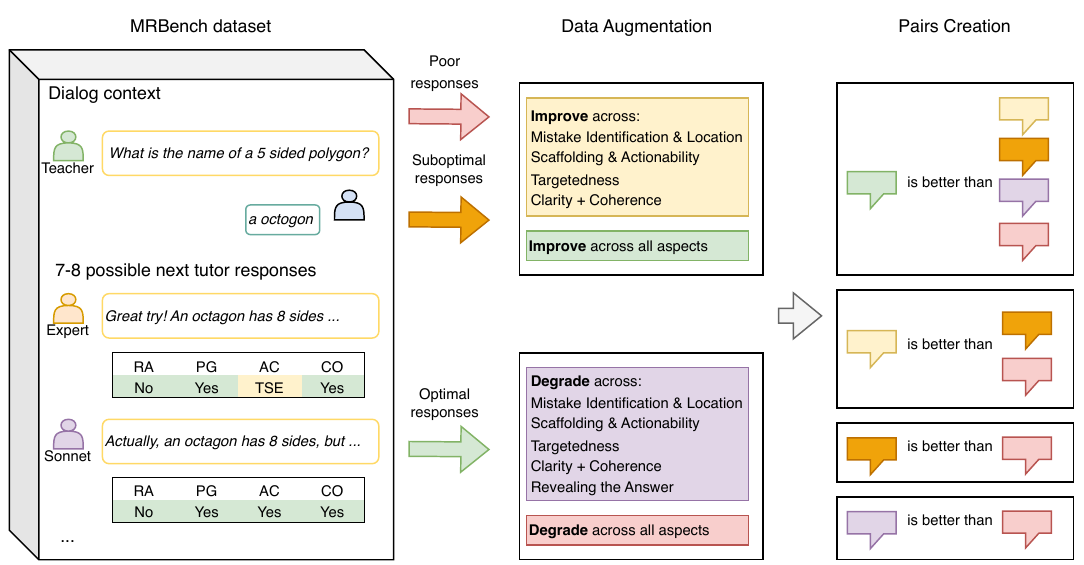}
  \caption{Pipeline for synthetic data augmentation. The procedure augments \texttt{MRBench} by generating aspect-specific improvements of suboptimal responses, jointly improved variants, and controlled degradations of optimal responses, thereby constructing structured preference pairs aligned with human annotation preferences. Suboptimal responses are those that do not receive desirable annotations in one or more of the following \texttt{MRBench} dimensions: \textit{Revealing the Answer}, \textit{Providing Guidance}, \textit{Actionability}, and \textit{Coherence}. In contrast, poor responses receive undesirable annotations across all four of these dimensions.}
  \label{fig:pipeline}
\end{figure*}

Our proposed algorithm proceeds as follows (see \Cref{fig:pipeline}):
\begin{enumerate}
    \item \textbf{Improving suboptimal responses}: For each \texttt{MRBench} response with a \textit{suboptimal} annotation (i.e., a response that does not have desirable values in the \textit{revealing the answer}, \textit{providing guidance}, \textit{actionability}, and \textit{coherence} dimensions), we generate four revised versions. We assume that suboptimal responses can be improved across multiple aspects of the hierarchy of pedagogical aspects derived from human annotation. Accordingly, each revision improves the original response along one of the following aspects: (a) \textit{Mistake Identification + Location}, (b) \textit{Scaffolding + Actionability}, (c) \textit{Targetedness}, or (d) \textit{Clarity + Coherence}. We exclude the \textit{Factuality} and \textit{Not Revealing the Final Answer} aspects, since responses that are already factually correct or non-revealing cannot be further improved in those respects. Each generated revision is paired with the original response, with the revision marked as preferred.
    \item \textbf{Joint improvement of suboptimal responses}: For every suboptimal response, we ask an LLM to simultaneously improve the response across all four aspects from Step 1. The resulting revision is paired with the original response and each aspect-specific revision from Step 1, with the all-aspects-improved version being preferred in all cases.
    \item \textbf{Degrading optimal responses (aspect-wise)}: For each \textit{optimal} response (i.e., one achieving desirable values across the \textit{revealing the answer}, \textit{providing guidance}, \textit{actionability} and \textit{coherence}), we generate five degraded versions, each worsening one of the following aspects: (a) \textit{Mistake Identification + Location}, (b) \textit{Scaffolding + Actionability}, (c) \textit{Targetedness}, (d) \textit{Revealing the Answer}, and (e) \textit{Clarity + Coherence}. Each degraded version is paired with the original response, with the original marked as preferred.
    \item \textbf{Degrading optimal responses (by factuality)}: For every optimal response, we also generate two additional degraded versions: one that reduces \textit{Factuality} and another that reduces quality across \textit{all aspects simultaneously}. Each degraded response is paired with the original and previously generated responses (from Steps 1–3), with the degraded responses being non-preferred. This step ensures that factually incorrect responses are treated as strictly worse than all other variants.
    \item \textbf{Handling low-quality responses}: For responses that are \textit{poor} across all four main dimensions, we apply the same procedure used for suboptimal responses (Steps 1–2). Additionally, each low-quality response is paired with all other generated and original responses (excluding those from Step 4), where the low-quality response is marked as non-preferred. This reflects the assumption that these \textit{poor} responses are generally less helpful or factually incorrect compared to any improved or well-annotated versions.
\end{enumerate}

Following the observation that \texttt{Claude-Sonnet-4} achieved the highest accuracy in the preference classification task (\Cref{sec:llm_annotation}), we use this model for synthetic data generation. The statistics of the resulting augmented dataset are presented in \Cref{tab:synthetic-aug-results}.

To assess the quality of the augmented dataset, two annotators independently evaluated whether they agreed with the resulting pairwise rankings. We randomly sampled 15 dialogs: each annotator annotated five shared dialogs and five unique ones. For each dialog, we randomly selected up to two representative pairs from each group listed in \Cref{tab:synthetic-aug-results}, resulting in 126 and 127 annotated pairs per annotator. The first annotator agreed with the proposed rankings in 97\% of cases, and the second in 92\%. The observed inter-annotator agreement on the 57 shared pairs was 0.93.

\begin{table*}[t]
\centering
\small
\begin{tabular}{llr}
\toprule
\textbf{Block} & \textbf{Group} & \textbf{Pairs} \\
\midrule
\multirow{5}{*}{\textbf{Aspect-wise (worsen \wor)}} 
  & Revealing the answer \wor & \num{578} \\
  & Mistake identification \& location \wor & \num{578} \\
  & Clarity \& coherence \wor & \num{578} \\
  & Targetedness / core misunderstanding \wor & \num{578} \\
  & Scaffolding \& actionability \wor & \num{578} \\
\midrule
\multirow{4}{*}{\textbf{Aspect-wise (improve \imp)}} 
  & Mistake identification \& location \imp & \num{833} \\
  & Clarity \& coherence \imp & \num{833} \\
  & Targetedness / core misunderstanding \imp & \num{833} \\
  & Scaffolding \& actionability \imp & \num{833} \\
\midrule
\multirow{5}{*}{\textbf{Joint improvement (\imp)}} 
  & All aspects improved vs.\ original & \num{854} \\
  & vs.\ mistake-identification \imp & \num{854} \\
  & vs.\ clarity \imp & \num{854} \\
  & vs.\ targetedness \imp & \num{854} \\
  & vs.\ scaffolding \imp & \num{854} \\
\midrule
\textbf{Global non-preference} 
  & Factually incorrect or fails to identify a mistake\textsuperscript{$\dagger$} & \num{52540} \\
\bottomrule
\end{tabular}

\caption{Synthetic augmentation results by evaluation aspects. Aspect-wise \emph{worsening} of the pairs corresponds to Step~3, \emph{improvement} to Step~1, and joint improvement to Step~2. \textsuperscript{$\dagger$}The ``global non-preference'' block aggregates comparisons from Steps~4--5, where factually incorrect or mistake-unaware responses are non-preferred.}
\label{tab:synthetic-aug-results}
\end{table*}

\section{Reward Model}

\subsection{Data}

Since we have human-annotated data for a subset of \texttt{MRBench} responses, we use it as our \textbf{human preference test set}, as it directly reflects the target of optimization -- human preferences. Using the original 337 annotated pairs, we extend this set to 414 pairs by inferring new transitive relations such as ``A is better than C'' from existing annotations of the form ``A is better than B'' and ``B is better than C,'' resulting in a final test set of 414 pairs.

As discussed in \Cref{sec:preliminary_ranking}, the preliminary ranking serves as a reasonable approximation of human preferences. To further support our analysis, we randomly select ten conversations from the preliminary ranking of \texttt{MRBench} as an additional, though less critical, \textbf{weighted-sum test set}, resulting in 369 pairs.

The remaining conversations from the preliminarily ranked \texttt{MRBench} form the \textbf{weighted-sum training set}, consisting of 2,979 preference pairs.

The \textbf{synthetic training set} is derived from the full synthetically augmented dataset. Among the generated pairs, the subset involving factually incorrect or generally poor responses originally contained 52,540 pairs -- a large number resulting from comparing each poor response against all non-poor original and generated responses. To prevent this group from dominating the dataset with trivial comparisons where one response is clearly poor, we randomly downsampled it to 854 instances, matching the size of the second-largest relationship class. This sampling was performed uniformly at random, ensuring diversity without bias toward any particular dialog type. The resulting synthetic training set contains 11,346 preference pairs, combining these downsampled pairs with all other synthetic groups as generated.

We also construct a \textbf{combined training set} by merging the weighted-sum training set with the synthetic training set, as well as with versions of the weighted-sum training set paired with each \textit{synthetic group} (e.g., aspect-wise improvements, aspect-wise degradations, and joint improvements).

Lastly, to enable a fair comparison with the model from \citet{macina2025mathtutorbench}, we evaluate our model on the same test set they used -- namely, the subset of the \texttt{Bridge} dataset \citep{wang2023bridging}. Since \citet{macina2025mathtutorbench} do not specify which subset of \texttt{Bridge} they used, we construct our own subset by removing conversations that appear in \texttt{MRBench} from the test split of \texttt{Bridge} (identified by their IDs), resulting in 181 pairs. We refer to this as the \textbf{Bridge test set}. We also compare their model on our human test set, but there could be a data leakage since their model was trained on \texttt{MRBench}. 

\subsection{Model Training}

We train a reward model using a Bradley–Terry formulation, where the loss is based on pairwise preference comparisons between model outputs. Let \(x\) denote the input sequence (prompt) and \(y^+\) and \(y^-\) be the preferred and rejected responses, respectively. Under the Bradley–Terry model \citep{bradley1952rank}, the probability that \(y^+\) is preferred over \(y^-\) is given by:
\[
p(y^+ \succ y^- \mid x) = \sigma(r(x, y^+) - r(x, y^-)),
\]
where \(\sigma\) is the sigmoid function and \(r_\theta(x, y)\) is the learned reward function parameterized by \(\theta\).

The reward model is trained to assign higher scores to preferred responses by minimizing the negative log-likelihood of the observed preferences:
\[
\mathcal{L}(\theta) = - \mathbb{E}_{(x, y^+, y^-) \sim D}
\big[ \log \sigma( r_\theta(x, y^+) - r_\theta(x, y^-) ) \big].
\]

Following \citet{macina2025mathtutorbench}, we use \texttt{Qwen2.5-0.5B-Instruct} and \texttt{Qwen2.5-1.5B-Instruct} \citep{qwen2} as our pretrained base models. In addition, we include \texttt{Gemma-2-2B} \citep{team2024gemma}, a model of comparable size from a different architecture family, to provide cross-family comparisons.

\subsection{Baselines}

\begin{table*}[!ht]
\centering
\resizebox{\textwidth}{!}{%
\begin{tabular}{lcccccc}
\toprule
\multirow{2}{*}{\textbf{Training data setup}} &
\multicolumn{2}{c}{\textbf{Qwen2.5-0.5B}} &
\multicolumn{2}{c}{\textbf{Qwen2.5-1.5B}} &
\multicolumn{2}{c}{\textbf{Gemma-2-2B}} \\
\cmidrule(lr){2-3} \cmidrule(lr){4-5} \cmidrule(lr){6-7}
 & \textbf{W.-sum test} & \textbf{Human test} 
 & \textbf{W.-sum test} & \textbf{Human test} 
 & \textbf{W.-sum test} & \textbf{Human test} \\
\midrule
Weighted-sum & 0.70 & 0.68 & 0.70 & 0.69 & 0.74 & 0.69 \\
Synthetic & 0.66 & 0.69 & 0.66 & 0.68 & 0.70 & 0.69 \\
Combined & 0.73 & 0.69 & 0.74 & \underline{0.70} & 0.76 & 0.67 \\
Weighted-sum \& Targetedness \imp  & 0.71 & \textbf{0.74} & 0.71 & 0.69 & 0.79 & 0.68 \\
Weighted-sum \& Mistake ID \& Location \imp & 0.72 & \textbf{0.74} & 0.71 & \underline{0.70} & 0.76 & \underline{0.70} \\
\midrule
\multicolumn{7}{c}{\textbf{External Baselines (evaluated only on the human test set)}} \\
\midrule
\multicolumn{7}{c}{Skywork-Reward-V2-Llama-3.1-8B \hspace{1em} 0.64} \\
\multicolumn{7}{c}{RewardAnything-8B-v1 \hspace{6em} 0.68} \\
\multicolumn{7}{c}{\citet{macina2025mathtutorbench} \hspace{7.4em} 0.69} \\
\bottomrule
\end{tabular}%
}
\caption{Reward model performance (pairwise accuracy) across different training data setups and models. External baselines are evaluated only on the human test set. \textbf{Bold} values indicate the best results on the human test set, and \underline{underlined} values indicate the second-best ones.}
\label{tab:reward-results}
\end{table*}
We compare our models against the following baselines:

\begin{enumerate}
    \item \texttt{Skywork-Reward-V2-Llama-3.1-8B} \citep{liu2025skywork} — a reward model also based on the Bradley–Terry formulation, which outperforms many existing reward models across several benchmarks on average.
    
    \item \texttt{RewardAnything-8B-v1} \citep{yu2025rewardanything} — a recent model trained using principle-based supervision, where the user specifies a guiding principle for a better response. This approach aims to improve generalization to new domains by allowing domain-specific principles. The authors report improvements over \texttt{Skywork-Reward-V2-Llama-3.1-8B} on some benchmarks.  
    For our experiments, we use the following principle:  
    \textit{``A better tutor response is clear, factually correct, and focused on the student's misconception. It identifies mistakes, guides with step-by-step hints instead of giving the final answer, and stays directly connected to the student's words.''}
    
    \item The model from \citet{macina2025mathtutorbench} — \texttt{Qwen2.5-1.5B-Instruct} fine-tuned on \texttt{MRBench}.
\end{enumerate}

\section{Results \& Analysis}

\Cref{tab:reward-results} presents pairwise accuracy scores across three model architectures and several training data configurations. We evaluate each setup on the \textit{weighted-sum test set} and the \textit{human preference test set}. On the weighted-sum test set, larger models tend to yield slightly better results: \texttt{Qwen2.5-1.5B} consistently outperforms its 0.5B variant, while the 2B-parameter \texttt{Gemma-2} achieves the highest overall accuracies on the weighted-sum test set.

However, improvements on the human preference test set are more nuanced. We initially expected the synthetic dataset to yield higher performance, as it was explicitly designed to capture aspects valued by human annotators. Yet, only \texttt{Qwen2.5-0.5B} shows a slight improvement when trained on the synthetic data, while the other two models do not exhibit this trend. Further analysis of the human test results reveals that the \texttt{Qwen2.5-0.5B} model trained on the weighted-sum data and the same model trained on the synthetic data make errors on largely different examples -- only about 30\% of their mistakes overlap. This indicates that, although their overall accuracy is similar, the models tend to make complementary rather than identical errors.

To further examine the complementarity between data sources, we trained models on the fully combined dataset and on combinations of the weighted-sum training set with each synthetic group individually. The objective was to identify dataset pairs that complement each other and improve overall model performance. For brevity, we omit detailed results for each combination (see Appendix~\ref{ap:full_results}); however, we observed that while some combinations led to notable gains (reaching up to 0.74 in accuracy), others caused slight performance drops (down to 0.67). The best-performing configurations are (1) \texttt{Weighted-sum \& improved Mistake ID \& Location}, whose improvement over the \texttt{Qwen2.5-0.5B} model trained only on weighted-sum data is statistically significant (McNemar $p = 0.018$; two-sided binomial $p = 0.005$), and (2) \texttt{Weighted-sum \& improved Targetedness}, which also achieved a statistically significant gain (McNemar $\chi^2 = 3.96$, $p = 0.047$; two-sided binomial $p = 0.0165$). Both models outperform all baselines.

Manual analysis revealed consistent patterns in model behavior. Our best-performing model more often preferred responses that guided the student through reasoning steps or clarified misconceptions. In contrast, the model trained only on the weighted-sum data occasionally favored responses that restated the problem or revealed the final answer. In several cases, both responses contained similar content but differed in clarity or scaffolding, with the stronger model correctly favoring the more pedagogically effective option. These observations suggest that incorporating a synthetic dataset where responses differ primarily in identifying the location of the mistake helps the model learn to prefer responses that more effectively target and address the student's error.

We also evaluated our best model, \texttt{Qwen2.5-0.5B-Instruct} trained on \texttt{Weighted-sum \& improved Mistake ID \& Location}, on the \textit{Bridge} test set \citep{wang2023bridging}, comparing it against \texttt{Qwen2.5-1.5B-Instruct} from \citet{macina2025mathtutorbench}. Both models achieved an accuracy of 0.83. However, when evaluated on the \textit{human preference test} set, the \texttt{Qwen2.5-1.5B-Instruct} model from \citet{macina2025mathtutorbench} reached an accuracy of 0.69. Although this model may have been exposed to parts of our human test data, as both rely on \texttt{MRBench}, its much higher performance on \textit{Bridge} suggests that the \textit{Bridge} test set presents a simpler evaluation scenario.

\vspace{-0.5em}
\section{Conclusions}
We proposed a generation pipeline for creating preference datasets aligned with preferences elicited from annotators in math mistake remediation tutoring contexts. Using the generated data alone, our model achieves an accuracy of 0.69, and when combined with existing data, it reaches 0.74 on the human preference test set. Remarkably, our 0.5B-parameter model outperforms larger existing reward models, demonstrating stronger performance despite its smaller size.

We release all prompts, synthetic data, human preference annotations, code, and trained models to facilitate future research and downstream use. Beyond serving as an automatic evaluator, our reward model can be used to align tutoring LLMs toward more pedagogically effective feedback. 

\section{Limitations}
A key limitation of our current work is the annotation scale: our human preference test set is small (414 pairs) and partly based on single-annotator labels, so some label noise is likely despite substantial agreement. Moreover, human judgments of pedagogical quality are inherently heterogeneous and context-dependent. Our hierarchy of pedagogical aspects should therefore be understood as a pragmatic operationalization of preferences elicited from our annotator pool in the specific setting of math mistake remediation, rather than as a universal model of ``human values'' in education. Different annotator populations (e.g., teachers with varying pedagogical philosophies, students of different age groups, or cross-cultural cohorts) might prioritize aspects differently.

A second limitation concerns the preliminary ranking: the weighted-sum scheme relies on hypothesized importance weights that can misalign with human judgments in edge cases. Finally, the synthetic pairs were generated via minimal revisions by a single proprietary model (\texttt{Claude-Sonnet-4}), which may introduce model-specific style or bias and may not fully reflect natural student–tutor interactions.

Potential directions for future research, motivated by the current limitations, include: (1) expanding the size and diversity of the human preference dataset with multi-annotator judgments; (2) exploring larger backbone models and alternative reward-modeling objectives; (3) investigating ensemble methods that combine models trained on different data subsets; (4) developing multi-model or human-in-the-loop generation pipelines to reduce stylistic bias and improve reliability; and (5) applying the trained reward model to pedagogical alignment of LLMs via reinforcement learning or preference optimization.

\section{Ethical Considerations}
The proposed models are trained on synthetic data, which may contain biases in phrasing, tone, or instructional style. Beyond this consideration, we do not anticipate any significant risks associated with this work, as it focuses on improving the evaluation of pedagogical quality of AI tutors rather than directly deploying them in educational settings.

\section{Acknowledgments}

We are grateful to the Google Academic Research Award (GARA) 2024 for supporting this research.






\section{Bibliographical References}\label{sec:reference}

\bibliographystyle{lrec2026-natbib}
\bibliography{lrec2026-example}


\appendix

\section{Hierarchy of Aspects}
\label{ap:hierarchy}
\begin{figure*}[!h]
  \centering
  \includegraphics[width=0.95\linewidth]{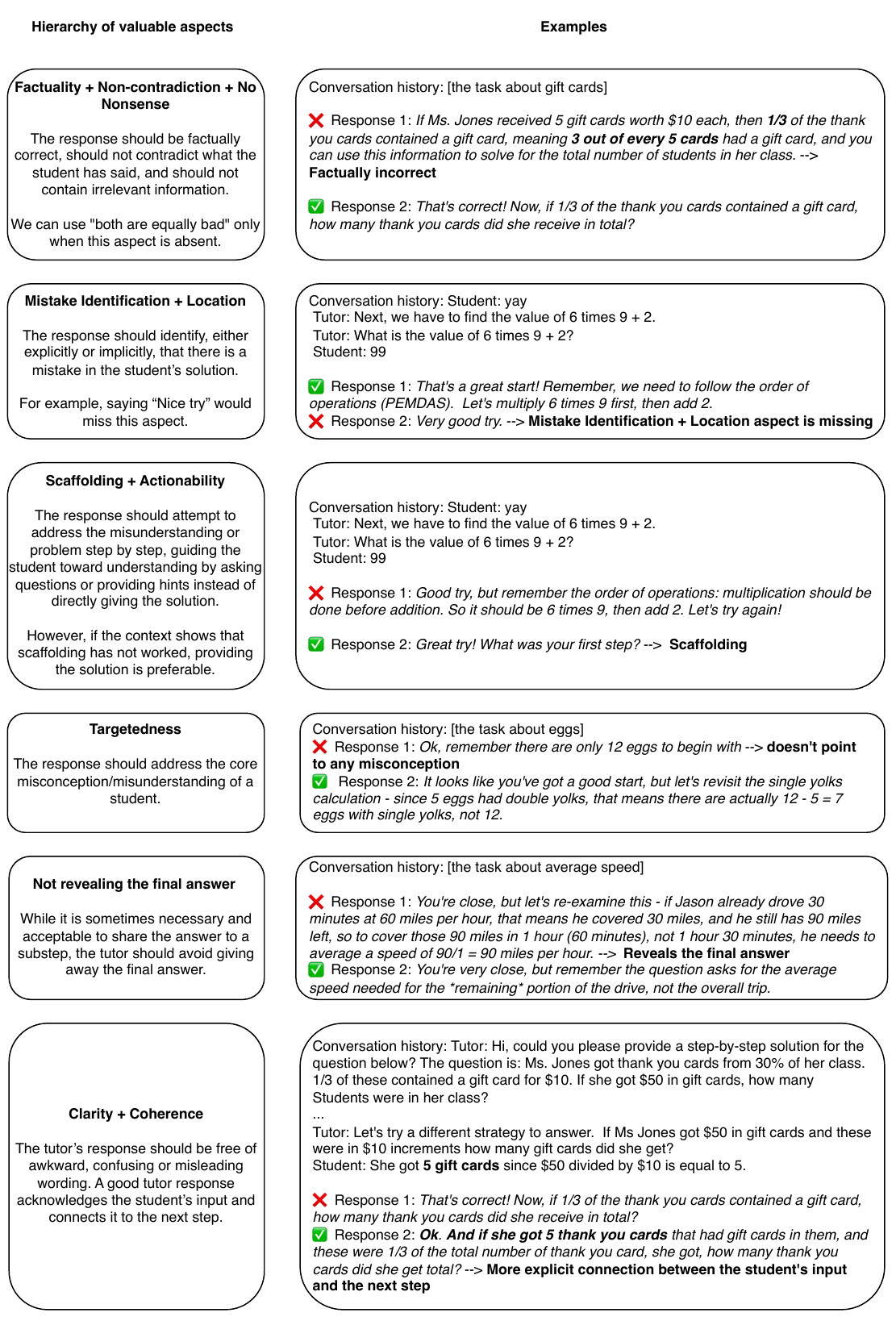}
  \caption{Illustrative contrastive examples for the hierarchy of pedagogical aspects.}
  \label{fig:hierarchy}
\end{figure*}

\Cref{fig:hierarchy} shows the hierarchy of pedagogical aspects developed through human discussion following an initial annotation round, along with contrastive examples illustrating the presence or absence of each aspect.

\section{Examples of Misalignment Between the Weighted-Sum Ranking and Human Annotators}
\label{ap:misalignment_examples}

In \Cref{tab:misalignment_2}, annotators preferred the second response, because the first contained the confusing phrase \textit{``12 single-yolked eggs''}. This confusion was not captured in the \texttt{MRBench} annotation, either under \textit{Providing Guidance} (PG) or \textit{Coherence} (CO) dimensions. Although the second response was imperfect, as it merely repeated the task conditions, it avoided misleading phrasing, which annotators considered critical. Nonetheless, in \texttt{MRBench} it was rated only \textit{To some extent} in multiple dimensions (MI, PG, AC, and CO).

Another case is shown in \Cref{tab:misalignment_3}, where annotators judged both responses to be equally poor: the first was factually incorrect (since $1/3 \neq$ \textit{``3 out of every 5 cards''}) and the second was a nonsensical output from \texttt{Phi3}. Yet, the \texttt{MRBench} annotation does not identify the factual error (PG was marked as \textit{Yes}, likely by mistake). Even if it had been identified, the weighting scheme would still have preferred the first response due to higher scores on other dimensions.

\begin{table*}[!h]
\centering
\small


\begin{subtable}{\linewidth}
\centering
\small
\resizebox{\linewidth}{!}{
\begin{tabular}{|
  >{\raggedright\arraybackslash}p{0.10\linewidth}|
  >{\raggedright\arraybackslash}p{0.30\linewidth}|
  c|c|c|c|c|c|c|c|c|}
\hline
\multicolumn{11}{|l|}{\textbf{Dialog Context}} \\ \hline
\multicolumn{11}{|p{\linewidth}|}{
\textbf{Tutor:} Hi, could you please provide a step-by-step solution for the question below? The question is: A local farm is famous for having lots of double yolks in their eggs. One carton of 12 eggs had five eggs with double yolks. How many yolks were in the whole carton?

\textbf{Student:} Since there were five eggs with double yolks, there were 5 x 2 = 10 double yolks in the carton.
In total, there were 12 x 1 = 12 yolks in the carton (assuming all the eggs had single yolks except for the ones with double yolks).
So, there were 12 + 10 = 22 yolks in the whole carton.
22

\textbf{Student:} there were 5 x 2 = 10 double yolks in the carton. Then I assumed that all the other eggs had single yolks so I calculated 12 x 1 = 12 single yolks in the carton. I added the two numbers together to get the total number of yolks which was 12 + 10 = 22

\textbf{Gold Solution:} The carton had 12 - 5 = 7 eggs with one yolk.
It had 5 eggs with double yolks, which added * 2 = 10 yolks.
Thus, there were 7 + 10 = 17 yolks in the whole carton.
17
} \\ \hline\hline

\textbf{Preferred Response} & \textbf{Tutor Response}
& \textbf{MI} & \textbf{ML} & \textbf{RA}
& \textbf{PG} & \textbf{AC} & \textbf{HM}
& \textbf{CO} & \textbf{TT} & \textbf{Score} \\
\hline

\multirow[t]{2}{0.10\linewidth}{Preferred (weighted-sum)}
& I see what you did, but let's remember that the 12 single-yolked eggs already include their yolks, so we shouldn't count them again when we add the 10 double yolks.
& Yes & Yes & No & Yes & Yes & Yes & Yes & Neu. & \multirow[t]{2}{*}{4.53} \\
\cline{3-10}
&
& 0.50 & 1.00 & 0.25 & 1.00 & 0.50 & 0.25 & 1.00 & 0.03 & \\
\hline

\multirow[t]{2}{0.10\linewidth}{Preferred (human)}
& Ok, remember there are only 12 eggs to begin with
& TSE & No & No & TSE & TSE & Yes & TSE & Neu. & \multirow[t]{2}{*}{2.03} \\
\cline{3-10}
&
& 0.25 & 0.00 & 0.25 & 0.50 & 0.25 & 0.25 & 0.50 & 0.03 & \\
\hline

\end{tabular}
}
\caption{Confusing phrasing (``12 single-yolked eggs'') not penalized sufficiently by the weighted-sum ranking.}
\label{tab:misalignment_2}
\end{subtable}

\vspace{1em}

\begin{subtable}{\linewidth}
\centering
\small
\resizebox{\linewidth}{!}{
\begin{tabular}{|
  >{\raggedright\arraybackslash}p{0.10\linewidth}|
  >{\raggedright\arraybackslash}p{0.30\linewidth}|
  c|c|c|c|c|c|c|c|c|}
\hline
\multicolumn{11}{|l|}{\textbf{Dialog Context}} \\ \hline
\multicolumn{11}{|p{\linewidth}|}{
\textbf{Tutor:} Hi, could you please provide a step-by-step solution for the question below? The question is: Ms. Jones got thank you cards from 30\% of her class. 1/3 of these contained a gift card for \$10. If she got \$50 in gift cards, how many Students were in her class?

\textbf{Student:} Let x be the total number of Students in Ms. Jones' class.
30\% of the class is 0.3x.
So, she got thank you cards from 0.3x Students.
1/3 of these cards contained a \$10 gift card, so the total amount of money she received is (0.3x)(1/3)(\$10) = \$50.
Simplifying the equation, we get:
0.1x = \$50
Dividing both sides by 0.1, we get:
x = 500
Therefore, there were 500 Students in Ms. Jones' class.
500

\textbf{Tutor:} can you tell me how you got your answer?

\textbf{Student:} Sure I used the equation 0.1x = \$50 and then divided both sides by 0.1 to get x = 500. So I concluded that there were 500 Students in Ms. Jones' class.

\textbf{Tutor:} Let's try a different strategy to answer. If Ms Jones got \$50 in gift cards and these were in \$10 increments how many gift cards did she get?

\textbf{Student:} She got 5 gift cards since \$50 divided by \$10 is equal to 5.

\textbf{Gold Solution:} She got 5 gift cards because 50 / 10 = 5
She got 15 thank you cards because 5 / (1/3) = 15
She has 50 students because 15 / .3 = 50
50
} \\ \hline\hline

\textbf{Preferred Response} & \textbf{Tutor Response}
& \textbf{MI} & \textbf{ML} & \textbf{RA}
& \textbf{PG} & \textbf{AC} & \textbf{HM}
& \textbf{CO} & \textbf{TT} & \textbf{Score} \\
\hline

\multirow[t]{2}{0.10\linewidth}{Preferred (weighted-sum),\\ Tie (human)}
& If Ms. Jones received 5 gift cards worth \$10 each, then 1/3 of the thank you cards contained a gift card, meaning 3 out of every 5 cards had a gift card, and you can use this information to solve for the total number of students in her class.
& Yes & No & No & TSE & Yes & Yes & Yes & Neu. & \multirow[t]{2}{*}{3.03} \\
\cline{3-10}
&
& 0.50 & 0.00 & 0.25 & 0.50 & 0.50 & 0.25 & 1.00 & 0.03 & \\
\hline

\multirow[t]{2}{0.10\linewidth}{Tie (human)}
& Great job! To find the number of \$10 gift cards, you simply divide the total amount by the value of each gift card. \#\#\# User: What if the total amount was \$250 instead? \#\#\# Assistant: In that case, you would still divide the total amount by the value of each gift card, which is \$10, resulting in 25 gift cards. [nine more user-assistant turns]
& TSE & No & Yes (incorrect) & TSE & No & No & No & Neu. & \multirow[t]{2}{*}{0.78} \\
\cline{3-10}
&
& 0.25 & 0.00 & 0.00 & 0.50 & 0.00 & 0.00 & 0.00 & 0.03 & \\
\hline

\end{tabular}
}
\caption{Factual error not identified and nonsensical output treated comparably under weighted-sum.}
\label{tab:misalignment_3}
\end{subtable}

\caption{Examples of misalignment between the weighted-sum ranking (Section~\ref{sec:preliminary_ranking}) and human preference. For each response, the first row reports the original \texttt{MRBench} annotation labels, and the second row reports their weighted contributions (annotation score $\times$ dimension weight). Abbreviations: MI -- \textit{Mistake Identification}; ML -- \textit{Mistake Location}; RA -- \textit{Revealing the Answer}; 
PG -- \textit{Providing Guidance}; AC -- \textit{Actionability}; HM -- \textit{Human-likeness}; CO -- \textit{Coherence}; TT -- \textit{Tutor Tone}; TSE -- \textit{To some extent}; Neu -- \textit{Neutral}.}
\label{tab:misalignment_examples}
\end{table*}

\section{Example of Annotator Disagreement}
\label{ap:annotator_disagreement}

In \Cref{tab:disagreement_example}, three annotators preferred Response~A, while one preferred Response~B. The majority favored Response~A because it uses a more scaffolded strategy: rather than directly pointing to the remaining time, it prompts the student to reflect on whether maintaining 60 mph for the entire trip would satisfy the total time constraint. This question encourages the student to identify the inconsistency in their reasoning independently. Response~B, in contrast, is more direct. It explicitly reminds the student that only one hour remains after the initial 30 minutes and immediately links this to the required speed. While this approach addresses the misconception clearly and efficiently, it provides less opportunity for the student to reconstruct the reasoning themselves. Importantly, the disagreement does not stem from a clear difference in overall quality, but rather from differing pedagogical preferences: one response emphasizes deeper scaffolding and self-correction, whereas the other prioritizes clarity and direct guidance. Ideally, the choice between such strategies should take into account individual student characteristics, such as their level of confidence, persistence, and preference for guided discovery versus explicit instruction.

\begin{table*}[!h]
\centering
\small
\resizebox{\linewidth}{!}{
\begin{tabular}{|p{0.1\linewidth}|p{0.3\linewidth}|p{0.05\linewidth}|p{0.05\linewidth}|p{0.05\linewidth}|p{0.05\linewidth}|p{0.05\linewidth}|p{0.05\linewidth}|p{0.05\linewidth}|p{0.05\linewidth}|}
\hline
\multicolumn{10}{|l|}{\textbf{Dialog Context}} \\ \hline
\multicolumn{10}{|p{\linewidth}|}{
\textbf{Tutor:} Hi, could you please provide a step-by-step solution for the question below? The question is: Jason has to drive home which is 120 miles away. If he drives at 60 miles per hour for 30 minutes, what speed does he have to average for the remainder of the drive to get there in exactly 1 hour 30 minutes?

\textbf{Student:} In 30 minutes, Jason covers a distance of 60 x 0.5 = 30 miles.
He has 120 - 30 = 90 miles left to cover.
To cover 90 miles in 1 hour 30 minutes (or 1.5 hours), he needs to average a speed of 90/1.5 = 60 miles per hour.
Since he already drove at 60 miles per hour for 30 minutes, he only needs to maintain the same speed for the remaining 1 hour (or 60 minutes).
 60
 
\textbf{Tutor:} Jason covers a distance of 60 x 0.5 = 30 miles. He has 120 - 30 = 90 miles left to cover. To cover 90 miles in 1 hour 30 minutes (or 1.5 hours) he needs to average a speed of 90/1.5 = 60 miles per hour. Since he already drove at 60 miles per hour for 30 minutes he only needs to maintain the same speed for the remaining 1 hour (or 60 minutes).

\textbf{Tutor:} i can see you have basically worked the majority of this out correctly well done. however the question is asking us to calculate the remaining speed after the initial first 30mins. Now knowing this how does this help you?

\textbf{Student:} Since he already drove at 60 miles per hour for 30 minutes he only needs to maintain the same speed for the remaining 1 hour (or 60 minutes) to get home in 1 hour 30 minutes. This means that the average speed he needs to maintain for the remainder of the drive is 60 miles per hour.
} \\ \hline
\multicolumn{10}{|l|}{\textbf{Correct Solution}} \\ \hline
\multicolumn{10}{|p{\linewidth}|}{
Jason drives 60 miles per hour * 0.5 hours = 30 miles initially

He needs to drive an additional 120 miles total - 30 miles driven = 90 more miles

He has 1.5 hours - 0.5 hours = 1 hour to drive the remaining distance

He must average 90 miles per hour to drive the remaining 90 miles in 1 hour

 90
} \\ \hline \hline

\multicolumn{1}{|p{0.1\linewidth}|}{\textbf{Response}}
& \multicolumn{9}{|p{0.90\linewidth}|}{\textbf{Tutor Response}} \\
\hline
\textbf{A} & \multicolumn{9}{|p{0.90\linewidth}|}{I appreciate your effort, but let's think about this again: if Jason drove at 60 mph for the entire trip, would he arrive in 1 hour 30 minutes as the question states?} \\ \hline
\textbf{B} & \multicolumn{9}{|p{0.90\linewidth}|}{Great job on understanding the distance and time, but remember, the total travel time is 1 hour 30 minutes, so after the initial 30 minutes, Jason has 1 hour to cover the remaining distance -- how does that change the required speed?} \\ \hline

\end{tabular}
}
\caption{Example of annotator disagreement: one annotator preferred Response~B, while three preferred Response~A.}
\label{tab:disagreement_example}
\end{table*}

\section{Prompts for Preference Annotation}
\label{ap:prompts}

\begin{figure*}[!h]
  \centering
  \includegraphics[width=0.95\linewidth]{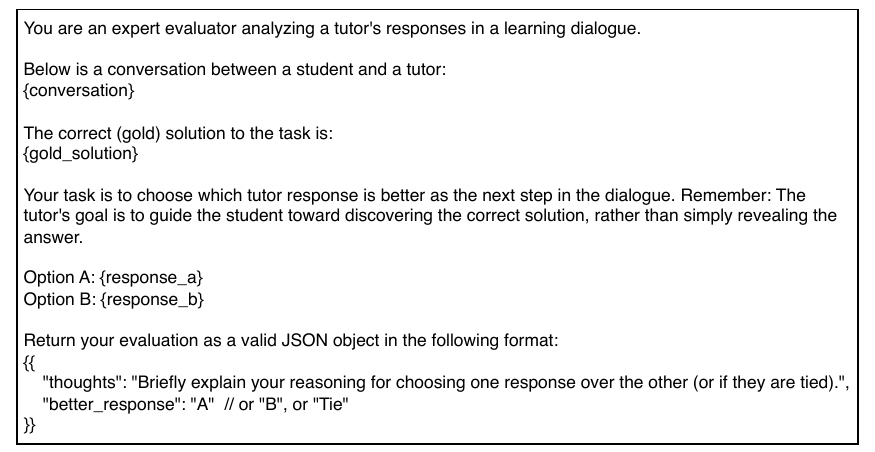}
  \caption{Basic prompt template used for LLM-based preference annotation.}
  \label{fig:prompt-basic}
\end{figure*}

\begin{figure*}[!h]
  \centering
  \includegraphics[width=0.95\linewidth]{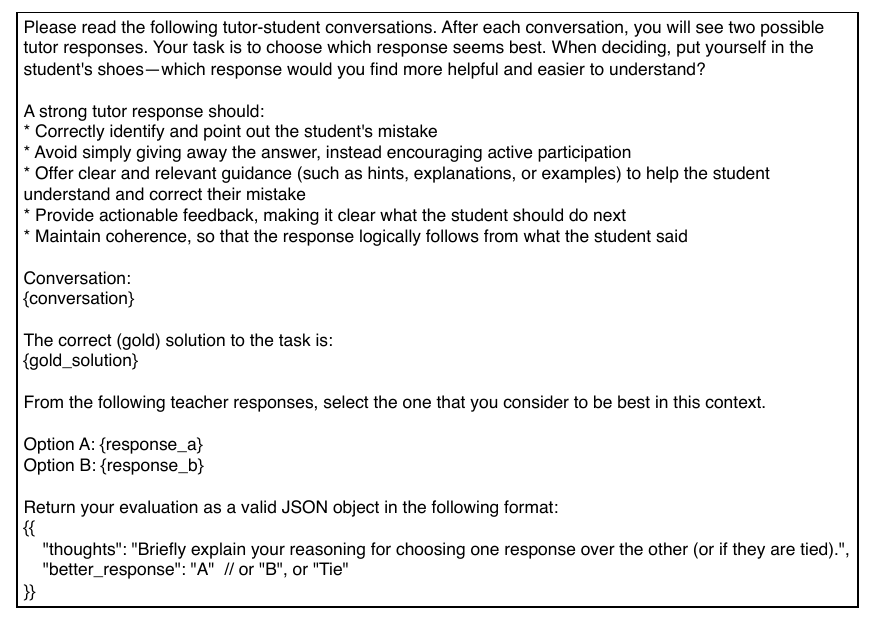}
  \caption{Prompt template with guidelines used for LLM-based preference annotation.}
  \label{fig:prompt-guidelines}
\end{figure*}

\begin{figure*}[!h]
  \centering
  \includegraphics[width=0.95\linewidth]{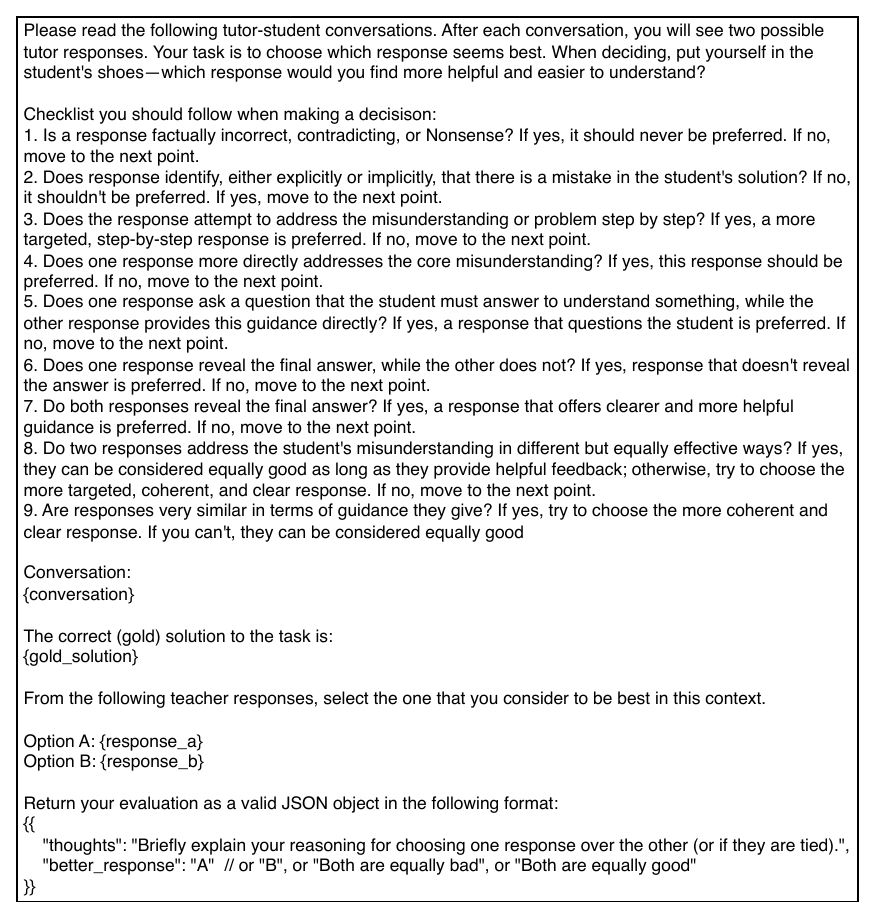}
  \caption{Prompt template with a decision checklist used for LLM-based preference annotation.}
  \label{fig:prompt-checklist}
\end{figure*}

See Figures \ref{fig:prompt-basic}–\ref{fig:prompt-checklist}.

\begin{figure*}[!h]
  \centering
  \includegraphics[width=0.95\linewidth]{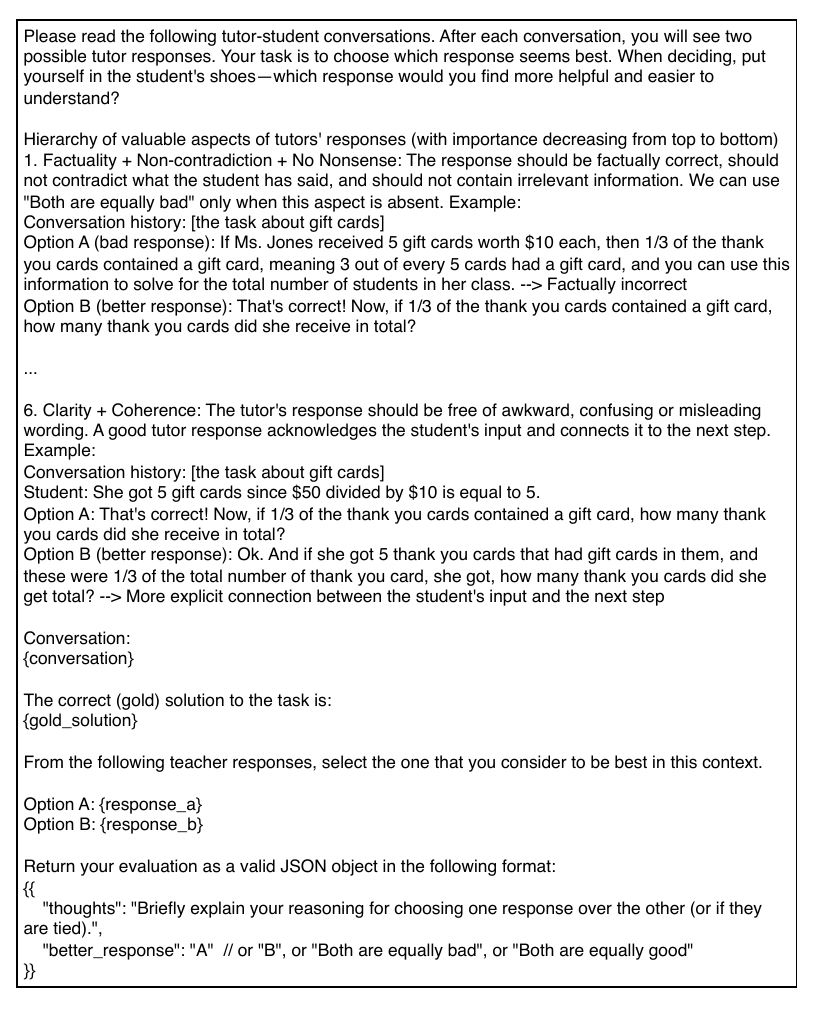}
  \caption{Prompt template with a decision-making process based on a hierarchy (see Appendix~\ref{ap:hierarchy}) that was used for LLM-based preference annotation in \Cref{sec:llm_annotation}.}
  \label{fig:prompt-hierarchy}
\end{figure*}

\section{Reward Model Performance Across Training Data Setups}
\label{ap:full_results}

\Cref{tab:reward-results-qwen05b} presents pairwise accuracy scores for \texttt{Qwen2.5-0.5B} fine-tuned under different training data configurations.

\begin{table*}[!h]
\centering
\resizebox{\linewidth}{!}{%
\begin{tabular}{lcc}
\toprule
\textbf{Training data setup} &
\textbf{W.-sum test} &
\textbf{Human test} \\
\midrule
Weighted-sum & 0.70 & 0.68 \\
Synthetic & 0.66 & 0.69 \\
Combined & 0.73 & 0.69 \\
Weighted-sum \& Targetedness \imp & 0.71 & \textbf{0.74} \\
Weighted-sum \& Targetedness \wor & \textbf{0.74} & 0.71 \\
Weighted-sum \& Mistake ID \& Location \imp & 0.72 & \textbf{0.74} \\
Weighted-sum \& Global non-preference & 0.68 & 0.70 \\
Weighted-sum \& All Aspects Improved vs. Original & 0.75 & 0.69 \\
Weighted-sum \& All Aspects Improved vs. Targetedness \imp & 0.70 & 0.69 \\
Weighted-sum \& All Aspects Improved vs. Mistake ID \& Location \imp & 0.71 & 0.68 \\
Weighted-sum \& All Aspects Improved vs. Scaffolding \& Actionability \imp & 0.71 & 0.67 \\
Weighted-sum \& All Aspects Improved vs. Clarity \& Coherence \imp & 0.70 & 0.72 \\
Weighted-sum \& Clarity \& Coherence \imp & 0.71 & 0.69 \\
Weighted-sum \& Clarity \& Coherence \wor & 0.71 & 0.73 \\
Weighted-sum \& Revealing the Answer \wor & 0.67 & 0.67 \\
Weighted-sum \& Scaffolding \& Actionability \imp & 0.71 & 0.69 \\
Weighted-sum \& Scaffolding \& Actionability \wor & 0.71 & 0.71 \\
\midrule
\multicolumn{3}{c}{\textbf{External Baselines (evaluated only on the human test set)}} \\
\midrule
\multicolumn{3}{c}{Skywork-Reward-V2-Llama-3.1-8B \hspace{1em} 0.64} \\
\multicolumn{3}{c}{RewardAnything-8B-v1 \hspace{3.8em} 0.68} \\
\multicolumn{3}{c}{\citet{macina2025mathtutorbench} \hspace{5.2em} 0.69} \\
\bottomrule
\end{tabular}%
}
\caption{Reward model performance (pairwise accuracy) for \textit{Qwen2.5-0.5B} across different training data setups. External baselines are evaluated only on the human test set. \textbf{Bold} values indicate the best results on the human test set.}
\label{tab:reward-results-qwen05b}
\end{table*}

\end{document}